\documentclass[letterpaper]{article} 
\usepackage{aaai2026}  
\usepackage{times}  
\usepackage{helvet}  
\usepackage{courier}  
\usepackage[hyphens]{url}  
\usepackage{graphicx} 
\usepackage{natbib}  
\usepackage{caption} 
\usepackage{algorithm}
\usepackage{algorithmic}
\usepackage{amsmath}
\usepackage{booktabs} 
\usepackage{multirow}

\usepackage{soul}
\usepackage{xcolor}
\usepackage{graphicx}
\usepackage{subcaption} 
\usepackage{amssymb}

\usepackage{newfloat}
\usepackage{listings}
\DeclareCaptionStyle{ruled}{labelfont=normalfont,labelsep=colon,strut=off} 
\floatstyle{ruled}
\newfloat{listing}{tb}{lst}{}
\floatname{listing}{Listing}
\title{Beyond Linear Surrogates: High-Fidelity Local Explanations for Black-Box Models}
\author{
    Sanjeev Shrestha,
    Rahul Dubey,
    Hui Liu
}

\affiliations{
    Department of Computer Science \\
    Missouri State University, USA\\
    \{ss472s, RahulDubey@, HuiLiu\}@MissouriState.edu
}
\usepackage{bibentry}

\begin{document}

\nocopyright
\maketitle

\begin{abstract}

With the increasing complexity of black-box machine learning models and their adoption in high-stakes areas, it is critical to provide explanations for their predictions. Existing local explanation methods lack in generating high-fidelity explanations. This paper proposes a novel local model agnostic explanation method to generate high-fidelity explanations using multivariate adaptive regression splines (MARS) and N-ball sampling strategies. MARS is used to model non-linear local boundaries that effectively captures the underlying behavior of the reference model, thereby enhancing the local fidelity. The N-ball sampling technique samples perturbed samples directly from a desired distribution instead of reweighting, leading to further improvement in the faithfulness. The performance of the proposed method was computed in terms of root mean squared error (RMSE) and evaluated on five different benchmark datasets with different kernel width. Experimental results show that the proposed method achieves higher local surrogate fidelity compared to baseline local explanation methods, with an average reduction of 32\% in root mean square error, indicating more accurate local approximations of the black-box model. Additionally, statistical analysis shows that across all benchmark datasets, the proposed approach results were statistically significantly better.  This paper advances the field of explainable AI by providing insights that can benefit the broader research and practitioner community.
\end{abstract}


\section{Introduction}
Artificial Intelligence (AI) has advanced modern information technologies and has become the core of intelligent systems. These AI-powered systems often operate autonomously, without direct human intervention in their design or deployment phases. As AI models increasingly influence critical aspects of human life across high-stake domains such as business, healthcare, manufacturing, and energy, there is a growing need to understand the decision-making process of these systems \cite{xai_survey}. However, it remains challenging to interpret complex Machine Learning (ML) models, despite their high predictive performance.

To address this challenge, eXplainable Artificial Intelligence (XAI) has emerged as a research field that deals with developing interpretable ML models and explanation techniques to foster transparency, trust, and accountability in AI systems \cite{xai}. 
Commonly, XAI approaches are classified into local vs global explanation methods. To locally explain a model's prediction, often a simple local surrogate model is used to mimic the behavior of a complex reference (true) model. 
The extent to which the surrogate approximates the behavior of the reference model is called \textit{fidelity}. 
A local surrogate focuses on a specific region of the model’s decision space, maintaining local fidelity while also preserving interpretability. Modern state-of-the-art techniques such as Local Interpretable Model-Agnostic Explanations (LIME) \cite{lime} adopt this approach by training a linear surrogate model on synthetic/perturbed samples labeled by the complex reference model, thereby generating explanations relevant to individual predictions.

Although methods like LIME have been widely adopted, prior studies have identified critical limitations, including instability, sensitivity to hyperparameter selection, and low fidelity in regions with nonlinear decision boundaries \cite{limeSurvey}. These challenges restrict the reliability of such explanations, particularly in high-stakes, safety-critical environments. The locality issue arises when the perturbed samples used to train the local surrogate model fail to adequately capture the true local decision boundary around the instance being explained. Similarly, the low fidelity issue, when the surrogate model is often constrained to be linear, prevents it from faithfully approximating the behaviour of complex black-box models, especially along highly nonlinear decision boundaries. Importantly, low fidelity does not arise only from poor locality caused by suboptimal sampling. It can also result from surrogate model that fails to capture non-linearity. \cite{lime_survey} Therefore, reliable local explanations require both sampling that maintains locality and a surrogate model capable of modeling nonlinear relationships.

In this study, we propose a novel high-fidelity local interpretable model agnostic explanation method (HF-LIME) that accurately learns the decision boundary localized to the instance being explained and models the nonlinear local classification boundaries. By relaxing assumptions of local linearity and feature independence, the proposed method aims to significantly enhance the local fidelity and interpretability of explanations for complex ML models, thereby contributing to more trustworthy and transparent decision-making systems. The proposed local model agnostic explanation method uses multivariate adaptive regression splines (MARS) and N-ball sampling strategies. MARS aims to capture reference model's non-linear local boundaries, thereby enhancing the local fidelity, and N-ball sampling technique samples perturbed samples directly from a desired distribution instead of reweighting, like in LIME, leading to further improvement in the fidelity.

To compare the performance of the proposed method against existing well-know local explanation method such as LIME and LEMON \cite{lemon}, we considered root mean squared error (RMSE) performance metric and evaluated on five different benchmark datasets. Prior research have shown that for local explanation, the width of neighborhood plays significant role in local surrogate model training \cite{locality}, and thus we compared the performance with different kernel widths. Experimental results show that the proposed method achieves higher local surrogate fidelity compared to baseline local explanation methods, with an average reduction of 32\% in RMSE, indicating more accurate local approximations of the black-box model. Additionally, statistical analysis shows that across all benchmark datasets, the proposed approach results were statistically significantly better.  This paper advances the field of explainable AI by providing insights that can benefit the broader research and practitioner community. The main contributions of this paper are as follows: 1) we introduce a novel local explanation framework that combines a sampling-based strategy with a piecewise linear surrogate model to improve local fidelity. 2) we present a comprehensive experimental evaluation on real-world datasets, including a systematic comparison with existing methods such as LIME and LEMON, to assess the effectiveness of the proposed approach.


\section{Related Work}
The LIME framework consists of two main components, a sampling method and the construction of a local surrogate model. Based on this, two research directions have been explored to improve LIME’s fidelity. The first focuses on replacing the linear local surrogate with a nonlinear model to achieve higher accuracy in representing local behavior. As LIME and most of its variants are based on linear surrogates, using nonlinear functions, such as quadratic models, can enhance fidelity, but may compromise interpretability \cite{qlime}. The second approach focuses on modifying the sampling strategy to better localize the decision boundary and enhance local fidelity.

Various studies have aimed to enhance the LIME sampling process to produce more relevant perturbations. LS-LIME \cite{lslime} focuses on sampling around the decision boundary to capture important regions better, while MPS-LIME \cite{mslime} accounts for feature dependencies to generate more realistic samples. CVAE-LIME \cite{cvaelime} aims to improve LIME’s local fidelity and interpretability by using Conditional Variational
Autoencoders (CVAE) for sample generation. LEMON \cite{lemon} introduced the N-ball sampling technique that uniformly samples within an N-ball for improved local approximation, and GLIME \cite{glime} employs a locally unbiased sampling strategy to enhance fidelity. US-LIME \cite{uslime} further improves robustness through a two-step uncertainty-based filtering that emphasizes samples near both the decision boundary and the instance of interest. However, these methods still rely on linear surrogates, limiting their ability to model complex nonlinear regions.

While sampling-based approaches aim to improve the quality and locality of perturbed samples, another line of research focuses on modifying the surrogate model itself. Several alternative models have been explored to replace the traditional linear approximation. LIME-SUP \cite{lime-sup} and LORE \cite{lore}  utilize tree-based techniques as replacement models. Sig-LIME \cite{sig-lime} integrates random forests to capture and utilize non-linear feature interactions, whereas LIMEtree \cite{limetree} adopts multi-output regression trees to achieve improved fidelity. However, tree-based models have lower predictive capabilities, which hinder their ability to approximate non-linear relationships accurately. Guo et al. proposed LEMNA \cite{lemna}, a high-fidelity explanation method for security applications that employs a Gaussian Mixture Model (GMM) to approximate nonlinear local boundaries and boost explanation fidelity. Inspired by this approach, Hung et al. introduced BMB-LIME \cite{bmb-lime}, which uses weighted multivariate adaptive regression splines (MARS) \cite{friedman1991multivariate} with bootstrap aggregating to enhance the local fidelity. Additionally, it employed a Bayesian framework of BayesLIME \cite{bayesLIME}  to model uncertainty and improve stability. Replacing the linear surrogate with a piecewise-linear model enhances LIME’s ability to approximate complex nonlinear decision boundaries, thereby improving explanation fidelity while maintaining interpretability. 

\section{Method}
\begin{figure*}[t]
    \centering
    \includegraphics[width=\textwidth]{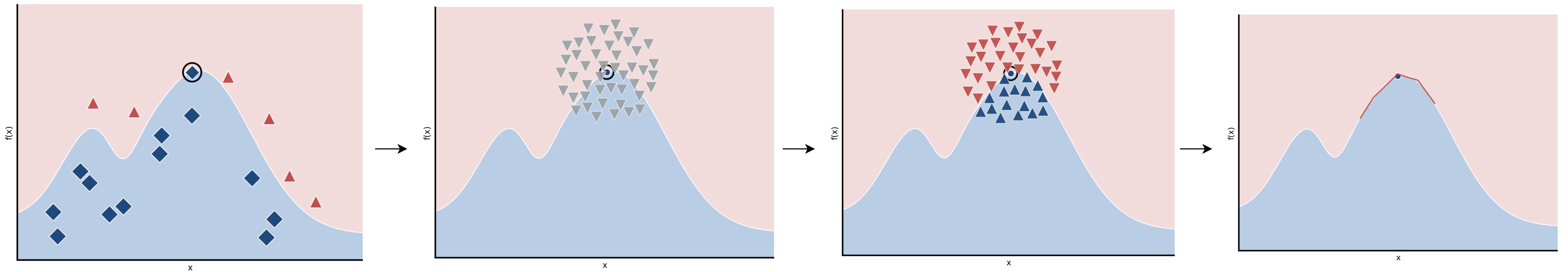}
   \caption{The workflow of the proposed approach includes the following steps: (1) first, select a target instance to be explained, (2) generate synthetic samples within its neighbourhood using N-ball sampling, (3) assign labels to these samples using the reference model, and (4) train a MARS surrogate model to approximate the non-linear local decision boundary. 
   Since the sampling method provides a denser set of local samples, the surrogate model captures the reference black box model's behavior more accurately.}
    \label{fig:pipeline}
\end{figure*}

This work proposes a novel and hybrid approach that incorporates a sampling-based technique and non-linear modeling of local decision boundaries, to enhance the local fidelity of the explanation method. The proposed method generates perturbed samples using the N-Ball sampling technique \cite{lemon}, which samples directly from the desired distribution, defined by a distance-kernel function, instead of reweighting samples. This creates a training dataset of closely related data points in the neighbourhood of the instance $x$ to be explained. 
Later, MARS~\cite{friedman1991multivariate} constructs a piecewise linear model suitable for high-dimensional problems.  Figure~\ref{fig:pipeline} depicts the sampling and non-linear model training.



\begin{algorithm}[t]
\caption{HF-LIME}
\label{algo:lemon_mars}
\begin{algorithmic}[1]

\STATE \textbf{Input:} Model $f$, number of perturbations $N$, instance $x \in \mathbb{R}^n$ ($n$ number of features), 
cumulative distribution function $F(\cdot)$, maximum radius $r_{\max}$, maximum degree of interaction $D$, 
maximum number of terms $Q$
\STATE \textbf{Output:} Feature contribution vector $\phi$

\STATE $\mathcal{Z} \leftarrow \{\}$ \hfill // initialize set of perturbed samples

\FOR{$i = 1$ \TO $N$}
    \STATE sample $y_i \sim \mathcal{N}(0, 1)$ $\in \mathbb{R}^n$
    \STATE $s_i \leftarrow y_i / \|y_i\|$ \hfill  // normalize
    
    \STATE sample $u_i \sim \mathcal{U}(0,1)$ $\in \mathbb{R}^n$
    \STATE $r_i \leftarrow F^{-1}(u_i)$  

    \STATE $z_i \leftarrow x + r_i \cdot s_i$ \hfill // N-ball sampling
    
    \STATE $\mathcal{Z} \leftarrow \mathcal{Z} \cup \{(z_i,\ f(z_i))\}$
\ENDFOR

\STATE $g \leftarrow \textsc{Surrogate}(\mathcal{Z}, D, Q)$ \hfill // $z$ as features, $f(z)$ as target
\STATE $\phi \leftarrow \textit{extract}(g)$ \hfill // extract the feature contributions

\STATE \textbf{return} $\phi$

\end{algorithmic}
\end{algorithm}


\subsection{N-ball Sampling}

In this method, synthetic data are generated within a unit hypersphere and then scaled by radius r (the region of interest), thereby translating them to be centred at $x$. LIME samples randomly from the entire feature space and then weights those samples based on their proximity to the instance $x$ being explained. However, this does not fully capture the true decision boundary in the local region around $x$, since training samples farther away are also considered. Instead, N-ball Sampling allows us to sample directly from the desired distribution around the local region of $x$. This generates samples that follow the dataset distribution around the instance x and attempts to localize the decision boundary.


Fishman~\cite{fishman} and Harman and Lacko~\cite{harman} present an efficient method for generating points uniformly within an $n$-dimensional hypersphere (i.e., an $n$-sphere). If $Y \sim \mathcal{N}(0,1)$, then the normalized vector $S_n = \frac{Y}{\|Y\|}$ is uniformly distributed on a unit $n$-sphere.
 By further scaling,
\begin{equation}
S_n \cdot U^{1/n}
\label{eq:unit_ball_sampling}
\end{equation}
where $U \sim \mathcal{U}(0,1)$, we obtain samples that are uniformly distributed within the unit $n$-ball, i.e., the region enclosed by the $n$-sphere.
This sampling approach ensures that all generated points lie strictly within the target region, bounded by a radius $r$ around $x$. As a result, a surrogate model can be trained on more relevant samples, thereby improving its ability to approximate the reference model's local behaviour. This can lead to more robust outcomes with reduced variance across algorithm runs.

Sampling is performed in two stages:
first, unit vectors are generated by sampling from a
Gaussian distribution and normalizing each vector to unit length. This ensures
points are uniformly distributed over the surface of a unit $n$-dimensional hypersphere in $\mathbb{R}^n$, where $n$ is the dimensionality of the feature space of instance $x$.
Second, radii are sampled according to the cumulative distribution function (CDF) weighted by distance kernel, and each vector is
scaled by its corresponding radius to obtain the perturbation within the $n$-sphere.
The cumulative distribution function for the radius of a sample is
\begin{equation}
\label{eq:cdf}
F(r) =
\frac{\int_0^r K(s) s^{n-1} ds}
{\int_0^{r_{\max}} K(s) s^{n-1} ds}, \quad r \le r_{\max}
\end{equation}
where $K(\cdot)$ is the distance kernel defined on the interval $[0, r_{\max}]$, where the maximum radius $r_{\max} > 0$ may vary depending on the kernel.
Sampling is done from Eq. \eqref{eq:cdf} using inverse transform sampling. Since the corresponding inverse CDF does not generally have a closed-form solution (e.g., for the Gaussian kernel used in LIME), it is approximated numerically.
Given a uniform random variable $u \sim \mathcal{U}(0,1)$, the sampled radius
$r$ is obtained as
\begin{equation}
\label{eq:r}
r = F^{-1}(u)
\end{equation}


When using a Gaussian distance kernel, the support of the kernel is unbounded,
which is incompatible with the bounded sampling region required for numerical
inverse transform sampling. To enable a fair comparison with the Gaussian kernel employed by LIME, following LEMON \cite{lemon}, we use the truncated Gaussian distance kernel, in which samples are drawn from a Gaussian distribution with the same kernel standard deviation and restricted to lie within $r_{\max}$. We choose $r_{\max}$
such that a fraction $p \in (0,1)$ of the sampled points lies within $r_{\max}$, and we set $p = 0.99$ for all experiments. $r_{\max}$ is defined as:
\begin{equation}
\label{eq:r_max}
r_{\max}
=
\sqrt{
2\sigma^2 \,
\Gamma^{-1}
\!\left(
\frac{n}{2},
(1 - p)\Gamma\!\left(\frac{n}{2}\right)
\right)
}
\end{equation}
where $\sigma$ is the kernel width, $n$ is the dimensionality of the instance feature space, $\Gamma(\cdot)$ denotes the gamma function, $\Gamma(z,s)=\int_s^{\infty} t^{z-1}e^{-t}\,dt$ is the upper incomplete gamma function, and $\Gamma^{-1}(\cdot,\cdot)$ is its inverse with respect to the second argument. 
Using a truncated Gaussian distance kernel with these parameters produces samples whose distributions closely align with LIME’s weighting scheme, thereby enabling a fair and consistent comparison between the two techniques.


\subsection{Modeling non-linearity with MARS}

After the sampling process, a local surrogate model is trained on a dataset
$\mathcal{Z} = \{(z_i, f(z_i))\}_{i=1}^{N}$,
where $z_i$ denotes a perturbed sample generated in the neighborhood of the instance $x$, and $f(z_i)$ represents the corresponding predictions of the black-box model $f$.
 In standard LIME, a linear regression model is typically used as the surrogate. In contrast, this work employs MARS to more accurately capture non-linear decision boundaries in the local neighborhood.

MARS attempts to capture nonlinearity by fitting piecewise linear regressions across different intervals of the input variable's range (i.e., subregions of the independent variable space). To achieve this, it partitions the input space into overlapping regions by keeping both the parent and its child basis functions after splitting a region. 
Each split generates a pair of hinge functions that model the behavior on either side of a selected knot. This removes discontinuities at partition boundaries, ensuring that the resulting approximations are continuous with continuous derivatives. This construction demonstrates significant flexibility, allowing it to capture highly nonlinear relationships with complex interactions. A stepwise regression approach is used to construct local surrogate model where a set of basis (hinge) functions and their interactions controls the prediction. The basis set $\mathcal{C}$ is defined as:
\begin{equation}
\mathcal{C} = 
\big\{(X_j - t)_+,\, (t - X_j)_+\big\}, \quad 
\begin{aligned}[t]
t &\in \{x_{1j}, \dots, x_{Nj}\},\\
j &= 1, \dots, P
\end{aligned}
\end{equation}
Here, $t$ represents all observed values across each predictor dimension. The hinge functions are formulated as:
$(x - t)_+ = x - t, \text{if } x > t$, otherwise 0, and 
$(t - x)_+ = t - x, \text{if } x < t$ otherwise 0, where the subscript ``$+$'' means the function takes only the positive value or is zero otherwise.
The computational complexity of fitting a MARS model is higher than that of a linear regression model, but it maintains about the same level of interpretability \cite{}, enabling greater fidelity.

A local surrogate MARS model is fitted on $\mathcal{Z}$ to approximate the local behavior of the black-box model $f$:
\begin{equation}
g(z) = \beta_0 + \sum_{q=1}^{Q} \beta_q \, h_q(z)
\label{eq:mars_model}
\end{equation}
where $h_q(z)$ is the $q^{th}$ basis function and its corresponding coefficient $\beta_q$ , $Q$ is the maximum number of basis functions. The contribution of each basis function at the explained instance $x$ is computed as \( \phi_q(x) = \beta_q \). Feature-level attributions are obtained by aggregating the contributions of all basis functions involving the same feature or feature interaction. 

\section{Experiment and Discussion}
\begin{table*}[ht!]
\centering
\caption{Comparing the average RMSE values obtained across all instances of datasets considering different black box models, explanation methods, and kernel widths. Best-performing scores are bolded.}
\label{tab:lime_lemon}
\renewcommand{\arraystretch}{1.3} %
\resizebox{1\textwidth}{!}{%
\begin{tabular}{l|c|r|r|r|r|r|r|r|r|r}

    \hline
    \multirow{2}{*}{Dataset}
      & \multirow{2}{*}{\shortstack{Kernel \\ width ($\sigma$)}}
      & \multicolumn{3}{c|}{Naive Bayes}
      & \multicolumn{3}{c|}{Random forest}
      & \multicolumn{3}{c}{MLP} \\
    \cline{3-11}
      &  & LIME & LEMON & HF-LIME & LIME & LEMON & HF-LIME & LIME & LEMON & HF-LIME \\
    \hline
    \multirow{6}{*}{\shortstack[l]{Wine \\($n=13$)}}
      & 0.1    & 0.009 & \textbf{0.003} & \textbf{0.003} & 0.041 & 0.018 & \textbf{0.005} & 0.064 & \textbf{0.007} & 0.008 \\
      & 0.3    & 0.044 & 0.026 & \textbf{0.024} & 0.118 & 0.051 & \textbf{0.017} & 0.101 & 0.040 & \textbf{0.019} \\
      & 0.5    & 0.103 & 0.071 & \textbf{0.065} & 0.185 & 0.082 & \textbf{0.030} & 0.219 & 0.142 & \textbf{0.051} \\
      & 1.0    & 0.258 & 0.224 & \textbf{0.188} & 0.156 & 0.120 & \textbf{0.051} & 0.262 & 0.205 & \textbf{0.106} \\
      & $\tfrac{3}{4}\sqrt{n}$
               & 0.652 & 0.303 & \textbf{0.245} & 0.376 & 0.124 & \textbf{0.065} & 0.499 & 0.273 & \textbf{0.202} \\
      & 4.0    & 0.848 & 0.282 & \textbf{0.226} & 0.545 & 0.120 & \textbf{0.065} & 0.811 & 0.339 & \textbf{0.203} \\
    \hline
    \multirow{6}{*}{\shortstack[l]{Diabetes\\($n=9$)}}
      & 0.1    & 0.018 & 0.016 & \textbf{0.002} & 0.072 & 0.036 & \textbf{0.009} & 0.018 & 0.015 & \textbf{0.001} \\
      & 0.3    & 0.057 & 0.031 & \textbf{0.014} & 0.141 & 0.053 & \textbf{0.023} & 0.050 & 0.026 & \textbf{0.006} \\
      & 0.5    & 0.080 & 0.046 & \textbf{0.034} & 0.112 & 0.064 & \textbf{0.036} & 0.068 & 0.031 & \textbf{0.017} \\
      & 1.0    & 0.120 & 0.110 & \textbf{0.098} & 0.105 & 0.088 & \textbf{0.054} & 0.068 & 0.061 & \textbf{0.055} \\
      & $\tfrac{3}{4}\sqrt{n}$
               & 0.387 & 0.257 & \textbf{0.212} & 0.247 & 0.100 & \textbf{0.062} & 0.242 & 0.145 & \textbf{0.139} \\
      & 4.0    & 0.687 & 0.348 & \textbf{0.261} & 0.419 & 0.096 & \textbf{0.060} & 0.465 & 0.192 & \textbf{0.185} \\
    \hline
    \multirow{6}{*}{\shortstack[l]{Breast cancer \\($n=32$)}}
      & 0.1    & 0.011 & 0.006 & \textbf{0.004} & 0.038 & 0.015 & \textbf{0.006} & 0.219 & 0.115 & \textbf{0.085} \\
      & 0.3    & 0.052 & 0.030 & \textbf{0.026} & 0.103 & 0.038 & \textbf{0.015} & 0.418 & 0.194 & \textbf{0.163} \\
      & 0.5    & 0.151 & 0.105 & \textbf{0.077} & 0.171 & 0.057 & \textbf{0.024} & 0.524 & 0.271 & \textbf{0.224} \\
      & 1.0    & 0.491 & 0.263 & \textbf{0.179} & 0.265 & 0.072 & \textbf{0.036} & 0.521 & 0.295 & \textbf{0.254} \\
      & $\tfrac{3}{4}\sqrt{n}$
               & 0.513 & \textbf{0.001} & \textbf{0.001} & 0.367 & 0.065 & \textbf{0.040} & 0.985 & \textbf{0.228} & 0.287 \\
      & 4.0    & 0.504 & \textbf{0.002} & \textbf{0.002} & 0.358 & 0.065 & \textbf{0.040} & 0.742 & 0.292 & \textbf{0.219} \\
    \hline
\multirow{6}{*}{\shortstack[l]{Credit \\($n=15$)}}
  & 0.1
    & 0.033 & 0.026 & \textbf{0.005}
    & 0.092 & 0.045 & \textbf{0.018}
    & 0.032 & 0.013 & \textbf{0.009} \\
  & 0.3
    & 0.218 & 0.172 & \textbf{0.063}
    & 0.150 & 0.056 & \textbf{0.033}
    & 0.153 & 0.084 & \textbf{0.075} \\
  & 0.5
    & 0.381 & 0.277 & \textbf{0.118}
    & 0.205 & 0.095 & \textbf{0.049}
    & 0.258 & 0.152 & \textbf{0.140} \\
  & 1.0
    & 0.333 & 0.304 & \textbf{0.180}
    & 0.169 & 0.150 & \textbf{0.074}
    & 0.269 & 0.250 & \textbf{0.234} \\
  & $\tfrac{3}{4}\sqrt{n}$
    & 0.405 & 0.237 & \textbf{0.196}
    & 0.320 & 0.145 & \textbf{0.090}
    & 0.650 & 0.321 & \textbf{0.308} \\
  & 4.0
    & 0.487 & 0.226 & \textbf{0.191}
    & 0.445 & 0.140 & \textbf{0.091}
    & 0.876 & 0.348 & \textbf{0.330} \\

    \hline
\multirow{6}{*}{\shortstack[l]{Liver \\($n=11$)}}
  & 0.1
    & 0.047 & 0.026 & \textbf{0.016}
    & 0.209 & 0.063 & \textbf{0.041}
    & 0.089 & 0.049 & \textbf{0.038} \\
  & 0.3
    & 0.449 & 0.295 & \textbf{0.147}
    & 0.257 & 0.081 & \textbf{0.055}
    & 0.233 & 0.184 & \textbf{0.154} \\
  & 0.5
    & 0.296 & 0.239 & \textbf{0.175}
    & 0.122 & 0.088 & \textbf{0.060}
    & 0.316 & 0.267 & \textbf{0.199} \\
  & 1.0
    & 0.088 & 0.079 & \textbf{0.075}
    & 0.104 & 0.093 & \textbf{0.065}
    & 0.285 & 0.266 & \textbf{0.220} \\
  & $\tfrac{3}{4}\sqrt{n}$
    & 0.025 & \textbf{0.014} & \textbf{0.014}
    & 0.187 & 0.093 & \textbf{0.068}
    & 0.471 & 0.297 & \textbf{0.269} \\
  & 4.0
    & 0.023 & \textbf{0.003} & \textbf{0.003}
    & 0.348 & 0.091 & \textbf{0.069}
    & 0.908 & 0.341 & \textbf{0.308} \\

\hline

    \end{tabular}
}
\end{table*}
In this section, we first define the evaluation metric for measuring the local fidelity of the explanation methods, and then present the experimental results.

\subsection{Evaluation Metric}

Local fidelity measures how well the local surrogate model
approximates the local behavior of the global black-box model. Here, RMSE is used to evaluate local fidelity. It indicates the deviation of the
predictions of the approximated model with respect to the reference model. A lower value of RMSE indicates higher fidelity, as it suggests that the local surrogate model is more accurate, being closer to the global model. RMSE is expressed as:

\begin{equation}
\text{RMSE}(\hat{y}^r, \hat{y}^s) = 
\sqrt{\frac{1}{m} \sum_{i=1}^m \left( \hat{y}^r_i - \hat{y}^s_i \right)^2 }
\end{equation}

where, $\hat{y}^r$ is the reference model, $\hat{y}^s$ is the surrogate model, and m is the synthetic sample in the area within radius $r_{\max}$ and an equivalent distance kernel to the ones used in LIME and LEMON. We generate $m = 50,000$ synthetic samples within
radius $r_{\max}$ using Eq. (3) and an equivalent distance kernel to the ones used in LIME and LEMON and compute the RMSE values between the prediction probabilities of the reference $\hat{y}^r$ and surrogate model $\hat{y}^s$ for all $m$ samples in the datasets.




\subsection{Experimental results}

We conducted experiments considering five benchmark datasets: Wine, Diabetes, Breast Cancer, Credit, Liver Patient \cite{lemon}. To explain the reference black-box models, we used three classifiers: a Gaussian Naive Bayes (GNB), a multilayer perceptron (MLP) with three hidden layers of 100 neurons each, and a Random Forest (RF) with 200 trees.  


To assess the robustness of the proposed method to the choice of locality parameter, we conduct experiments across a wide range of kernel width values, $\sigma \in \{0.1, 0.3, 0.5, 1.0, 4.0,$ and $\tfrac{3}{4}\sqrt{n}$\}, where $n$ is the input dimensionality. The default kernel width in LIME is set to $\tfrac{3}{4}\sqrt{n}$; however, due to its relatively large magnitude ($>1$ for $n\geq2$), it cannot be regarded as truly local. Finally, $r_{\text{max}}$ was calculated using Eq. \eqref{eq:r_max} with $p = 0.999$.

\begin{figure*}[t]
\centering 
\begin{subfigure}[t]{0.3\textwidth}
  \centering
  \includegraphics[width=\linewidth]{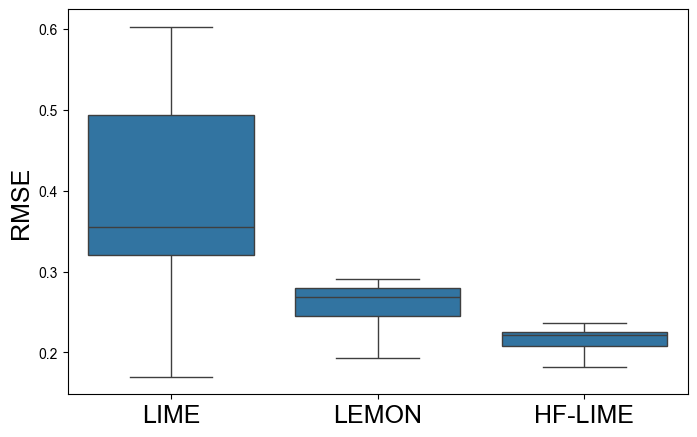}
  \caption{Diabetes (GNB)}
\end{subfigure}\hfill
\begin{subfigure}[t]{0.3\textwidth}
  \centering
  \includegraphics[width=\linewidth]{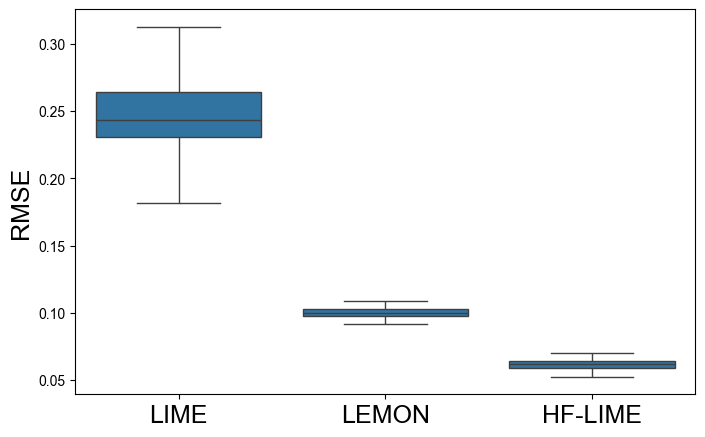}
  \caption{Diabetes (RF)}
\end{subfigure}\hfill
\begin{subfigure}[t]{0.3\textwidth}
  \centering
  \includegraphics[width=\linewidth]{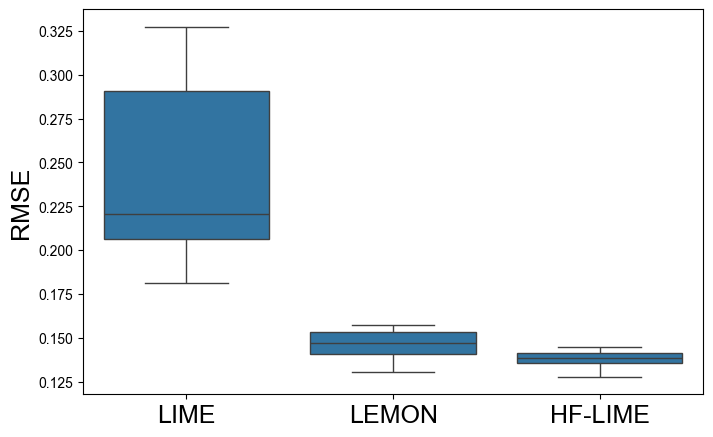}
  \caption{Diabetes (MLP)}
\end{subfigure}

\vspace{0.6em}

\begin{subfigure}[t]{0.3\textwidth}
  \centering
  \includegraphics[width=\linewidth]{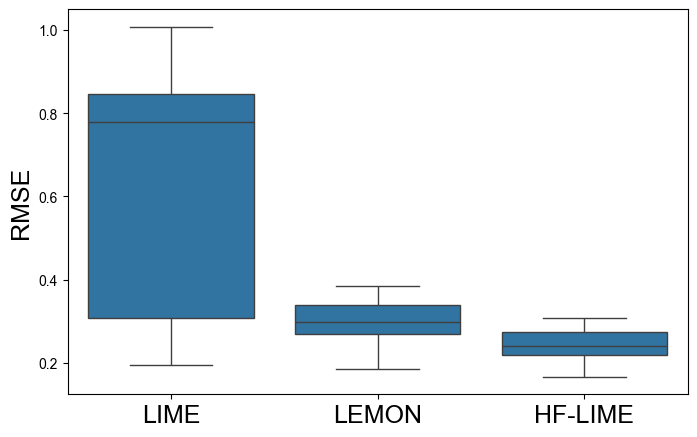}
  \caption{Wine (GNB)}
\end{subfigure}\hfill
\begin{subfigure}[t]{0.3\textwidth}
  \centering
  \includegraphics[width=\linewidth]{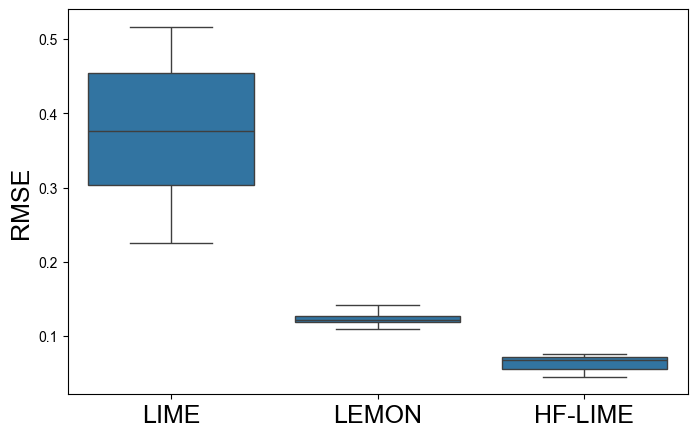}
  \caption{Wine (RF)}
\end{subfigure}\hfill
\begin{subfigure}[t]{0.3\textwidth}
  \centering
  \includegraphics[width=\linewidth]{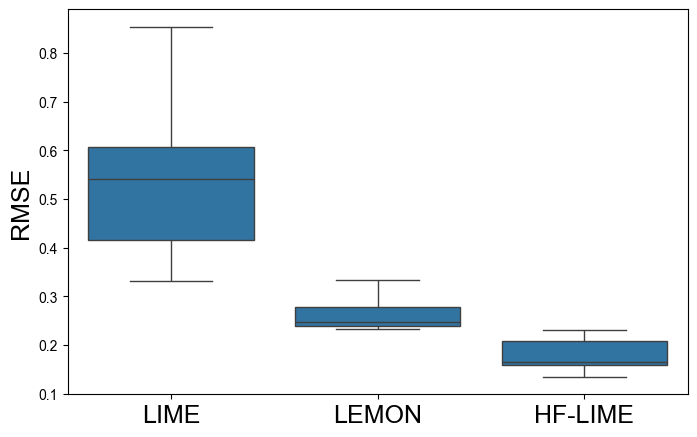}
  \caption{Wine (MLP)}
\end{subfigure}

\caption{RMSE distributions measuring the local fidelity of LIME, LEMON, and HF-LIME methods across GNB, RF, and MLP classifiers at the default kernel width $\sigma=(3/4)\sqrt{n}$ on the Diabetes and Wine datasets. Lower RMSE indicates higher local fidelity. HF-LIME consistently achieves lower error and reduced variance across classifiers and datasets, indicating improved stability of explanations.}
\label{fig:boxplot}
\end{figure*}

Table \ref{tab:lime_lemon} shows the comparison of the average RMSE scores across the datasets between HF-LIME and the baselines (LIME and LEMON). Bolded values are the best-performing scores. Results show that HF-LIME consistently achieves the lowest RMSE values across all datasets, models, and kernel widths. In particular, HF-LIME improves the baseline LIME method while also outperforming LEMON in most cases. On average, HF-LIME achieves a 64.8\% reduction in RMSE compared to LIME, while LEMON achieves a 47.03\% reduction again compared to LIME, indicating that HF-LIME provides significantly higher local fidelity than both LIME and LEMON. Furthermore, HF-LIME achieves a 32.03\% lower RMSE compared to LEMON, demonstrating its superior performance among baselines. 

On Wine dataset, HF-LIME provides significant improvements, particularly at larger kernel widths (e.g., at $\sigma=4.0$)
The Diabetes dataset, which has lower dimensionality, also benefits significantly: at $\sigma=0.1$, RF RMSE drops from 0.072 (LIME) to 0.009 (HF-LIME). This effect becomes more evident in the Breast Cancer dataset, which is higher-dimensional ($n=32$), where LIME produces unstable results with higher error values. (e.g., MLP RMSE of 0.985 at $\sigma=(3/4)\sqrt{n}$), while HF-LIME maintains consistently lower values (0.287 under the same setting). Although LEMON occasionally achieves low RMSE (e.g., GNB at $\sigma=(3/4)\sqrt{n}$), these results are not consistently observed across classifiers.

The choice of kernel width significantly affects local fidelity, as it defines the extend of the local neighbourhood. While LIME’s performance decreases rapidly as $\sigma$ increases, both LEMON and HF-LIME show greater flexibility. 
Overall, HF-LIME provides the most stable local explanations, consistently maintaining lower RMSE across all kernel widths.


\begin{table*}[ht!]
\centering
\caption{Standard deviation of RMSE values, averaged across all kernel widths. Best-performing (lowest) scores are bolded. HF-LIME performed better than others by having a smaller standard deviation.}
\label{tab:lime_lemon_avg_std}

\renewcommand{\arraystretch}{1.3}
\begin{tabular}{l|r|r|r|r|r|r|r|r|r}
\hline
\multirow{2}{*}{Dataset}
& \multicolumn{3}{c|}{Naive Bayes}
& \multicolumn{3}{c|}{Random Forest}
& \multicolumn{3}{c}{MLP} \\
\cline{2-10}
& LIME & LEMON & HF-LIME 

  & LIME & LEMON & HF-LIME
  & LIME & LEMON & HF-LIME \\
\hline

Wine

& 0.269 & 0.040 & \textbf{0.031}
& 0.077 & 0.008 & \textbf{0.008}
& 0.118 & 0.036 & \textbf{0.027} \\

Diabetes

&  0.108 & 0.032 & \textbf{0.021}
&  0.031 & \textbf{0.004} & \textbf{0.004}
&  0.044 & 0.007 & \textbf{0.004} \\

Breast Cancer

& 0.384 & \textbf{0.002} & \textbf{0.002}
&  0.034 & \textbf{0.002} & \textbf{0.002}
&  0.168 & 0.011 & \textbf{ 0.009} \\

Credit

&  0.164 & 0.090 & \textbf{ 0.036}
& 0.060 & 0.014 & \textbf{ 0.008}
& 0.125 & 0.034 & \textbf{0.031} \\

Liver

& 0.096 &  0.063 & \textbf{ 0.038}
&  0.059 & \textbf{0.012} & 0.013
&  0.101 & 0.053 & \textbf{ 0.046} \\
\hline

\end{tabular}
\end{table*}

Fig. \ref{fig:boxplot} illustrates the distribution of RMSE values computed for LIME, LEMON, and HF-LIME across GNB, RF, and MLP classifiers for Diabetes and Wine datasets. We kept the default kernel width to $\sigma = (3/4)\sqrt{n}$, as used by LIME for a fair comparison. Across all settings, HF-LIME consistently achieves the lowest median RMSE and a substantially smaller interquartile range than both LIME and LEMON. This indicates that the proposed method produces explanations that more closely approximate the behavior of the underlying black-box model while also exhibiting improved stability across instances. In contrast, LIME shows higher RMSE and greater variance, suggesting low fidelity and more unstable local explanations.
To assess whether the observed differences in RMSE across explanation methods are statistically significant, we conducted a one-tailed paired t-test on the RMSE values calculated for all instances of the datasets. Across all datasets and classifiers, HF-LIME achieves statistically significant reductions in RMSE compared to LIME and LEMON ($p < 0.05$), validating the effectiveness of the proposed approach in improving local fidelity. 

Figure~\ref{fig:boxplot} also suggests that the spread of RMSE values is the smallest using the proposed method compared to the other two approaches. Thus, to measure the stability of local surrogate fidelity, we computed the standard deviation of RMSE values across all datasets and kernel widths. 
Table \ref{tab:lime_lemon_avg_std} shows the average standard deviation, obtained by first computing the standard deviation of RMSE values across instances for each kernel width and then averaging these values across all kernel widths. This result indicates that the proposed approach not only generates high-fidelity local explanations but is also stable, with lower variance than the reference model. 


Overall, these experimental results indicate that HF-LIME offers high-fidelity, robust, and stable explanations compared to both LIME and LEMON. The boost in fidelity comes from integrating N-ball sampling, which enhances the locality of the decision boundary, with the MARS surrogate model, which effectively captures the underlying nonlinear relationships. Its stability across kernel widths, effectiveness on higher-dimensional datasets, and consistent improvements across classifiers demonstrate its suitability as a more accurate method for explaining reference black-box models.

\section{Conclusion}
This paper proposes a novel method designed to improve the fidelity of local explanation.  By integrating N-ball sampling with MARS surrogate, it localizes the decision boundary and captures nonlinear relationships more effectively than existing methods such as LIME and LEMON. Through experiments on five real-world datasets, we demonstrated that HF-LIME consistently outperforms baseline methods in terms of RMSE. On average, HF-LIME achieves a 64.8\% reduction in RMSE compared to LIME and a 32\% reduction compared to LEMON, confirming its ability to generate higher-fidelity explanations. The statistical analysis shows that HF-LIME achieves statistically significant improvements in local fidelity while demonstrating lower variability across kernel widths, resulting in more stable local explanations than existing methods. Future work will focus on further improving explanation stability, consistency, and extending the proposed method to other data modalities, such as image and time-series data. Addressing these prospects will further advance the goal of reliable, faithful, and robust explanations for complex real-world systems.



\bibliography{ref}

@article{xai,
title = {Explainable Artificial Intelligence (XAI): What we know and what is left to attain Trustworthy Artificial Intelligence},
journal = {Information Fusion},
volume = {99},
pages = {101805},
year = {2023},
issn = {1566-2535},
doi = {https://doi.org/10.1016/j.inffus.2023.101805},
url = {https://www.sciencedirect.com/science/article/pii/S1566253523001148},
author = {Sajid Ali and Tamer Abuhmed and Shaker El-Sappagh and Khan Muhammad and Jose M. Alonso-Moral and Roberto Confalonieri and Riccardo Guidotti and Javier {Del Ser} and Natalia Díaz-Rodríguez and Francisco Herrera}
}

@inproceedings{limeSurvey,
        title={Which LIME should I trust? Concepts, Challenges, and Solutions},
        author={Patrick Knab and Sascha Marton and Udo Schlegel and Christian Bartelt},
        booktitle={Proceedings of the XAI 2025 Conference},
        year={2025},
        eprint={2503.24365},
        archivePrefix={arXiv},
        primaryClass={cs.LG},
        url={https://arxiv.org/abs/2503.24365},
    }

@article{bmb-lime,
title = {BMB-LIME: LIME with modeling local nonlinearity and uncertainty in explainability},
journal = {Knowledge-Based Systems},
volume = {294},
pages = {111732},
year = {2024},
issn = {0950-7051},
doi = {https://doi.org/10.1016/j.knosys.2024.111732},
url = {https://www.sciencedirect.com/science/article/pii/S0950705124003678},
author = {Yu-Hsin Hung and Chia-Yen Lee},
}

@inproceedings{lemna,
author = {Guo, Wenbo and Mu, Dongliang and Xu, Jun and Su, Purui and Wang, Gang and Xing, Xinyu},
title = {LEMNA: Explaining Deep Learning based Security Applications},
year = {2018},
isbn = {9781450356930},
publisher = {Association for Computing Machinery},
address = {New York, NY, USA},
url = {https://doi.org/10.1145/3243734.3243792},
doi = {10.1145/3243734.3243792},
booktitle = {Proceedings of the 2018 ACM SIGSAC Conference on Computer and Communications Security},
pages = {364–379},
numpages = {16},
location = {Toronto, Canada},
series = {CCS '18}
}

@InProceedings{lemon,
author="Collaris, Dennis
and Gajane, Pratik
and Jorritsma, Joost
and van Wijk, Jarke J.
and Pechenizkiy, Mykola",
editor="Cr{\'e}milleux, Bruno
and Hess, Sibylle
and Nijssen, Siegfried",
title="LEMON: Alternative Sampling for More Faithful Explanation Through Local Surrogate Models",
booktitle="Advances in Intelligent Data Analysis XXI",
year="2023",
publisher="Springer Nature Switzerland",
address="Cham",
pages="77--90",
isbn="978-3-031-30047-9"
}

@ARTICLE{qlime,
  title    = "{QLIME-A} quadratic Local Interpretable Model-agnostic
              explanation approach",
  author   = "Bramhall, Steven and Horn, Hayley and Tieu, Michael and Lohia,
              Nibhrat",
  journal  = "SMU Data Science Review",
  volume   =  3,
  number   =  1,
  pages    = "4",
  year     =  2020
}

@misc{locality,
      title={Defining Locality for Surrogates in Post-hoc Interpretablity}, 
      author={Thibault Laugel and Xavier Renard and Marie-Jeanne Lesot and Christophe Marsala and Marcin Detyniecki},
      year={2018},
      eprint={1806.07498},
      archivePrefix={arXiv},
      primaryClass={cs.LG},
      url={https://arxiv.org/abs/1806.07498}, 
}

@article{lslime,
  author       = {Thibault Laugel and
                  Xavier Renard and
                  Marie{-}Jeanne Lesot and
                  Christophe Marsala and
                  Marcin Detyniecki},
  title        = {Defining Locality for Surrogates in Post-hoc Interpretablity},
  journal      = {CoRR},
  volume       = {abs/1806.07498},
  year         = {2018},
  url          = {http://arxiv.org/abs/1806.07498},
  eprinttype    = {arXiv},
  eprint       = {1806.07498},
  bibsource    = {dblp computer science bibliography, https://dblp.org}
}

@misc{mslime,
      title={A Modified Perturbed Sampling Method for Local Interpretable Model-agnostic Explanation}, 
      author={Sheng Shi and Xinfeng Zhang and Wei Fan},
      year={2020},
      eprint={2002.07434},
      archivePrefix={arXiv},
      primaryClass={cs.LG},
      url={https://arxiv.org/abs/2002.07434}, 
}

@article{uslime,
title = {US-LIME: Increasing fidelity in LIME using uncertainty sampling on tabular data},
journal = {Neurocomputing},
volume = {597},
pages = {127969},
year = {2024},
issn = {0925-2312},
doi = {https://doi.org/10.1016/j.neucom.2024.127969},
url = {https://www.sciencedirect.com/science/article/pii/S0925231224007409},
author = {Hamid Saadatfar and Zeinab Kiani-Zadegan and Benyamin Ghahremani-Nezhad},
}

@misc{glime,
      title={GLIME: General, Stable and Local LIME Explanation}, 
      author={Zeren Tan and Yang Tian and Jian Li},
      year={2023},
      eprint={2311.15722},
      archivePrefix={arXiv},
      primaryClass={cs.LG},
      url={https://arxiv.org/abs/2311.15722}, 
}

@ARTICLE{cvaelime,
  author={Yasui, Daisuke and Sato, Hiroshi},
  journal={IEEE Access}, 
  title={Improving Local Fidelity and Interpretability of LIME by Replacing Only the Sampling Process With CVAE}, 
  year={2025},
  volume={13},
  number={},
  pages={53084-53099},
  doi={10.1109/ACCESS.2025.3553505}
}

@misc{lime-sup,
      title={Locally Interpretable Models and Effects based on Supervised Partitioning (LIME-SUP)}, 
      author={Linwei Hu and Jie Chen and Vijayan N. Nair and Agus Sudjianto},
      year={2018},
      eprint={1806.00663},
      archivePrefix={arXiv},
      primaryClass={stat.ML},
      url={https://arxiv.org/abs/1806.00663}, 
}

@misc{lore,
      title={Local Rule-Based Explanations of Black Box Decision Systems}, 
      author={Riccardo Guidotti and Anna Monreale and Salvatore Ruggieri and Dino Pedreschi and Franco Turini and Fosca Giannotti},
      year={2018},
      eprint={1805.10820},
      archivePrefix={arXiv},
      primaryClass={cs.AI},
      url={https://arxiv.org/abs/1805.10820}, 
}

@Article{limetree,
AUTHOR = {Sokol, Kacper and Flach, Peter},
TITLE = {LIMETREE: Consistent and Faithful Surrogate Explanations of Multiple Classes},
JOURNAL = {Electronics},
VOLUME = {14},
YEAR = {2025},
NUMBER = {5},
ARTICLE-NUMBER = {929},
URL = {https://www.mdpi.com/2079-9292/14/5/929},
ISSN = {2079-9292},
DOI = {10.3390/electronics14050929}
}

@ARTICLE{sig-lime,
  author={Abdullah, Talal Ali Ahmed and Zahid, Mohd Soperi Mohd and Turki, Ahmad F. and Ali, Waleed and Jiman, Ahmad A. and Abdulaal, Mohammed J. and Sobahi, Nebras M. and Attar, Eyad T.},
  journal={IEEE Access}, 
  title={Sig-Lime: A Signal-Based Enhancement of Lime Explanation Technique}, 
  year={2024},
  volume={12},
  number={},
  pages={52641-52658},
  doi={10.1109/ACCESS.2024.3384277}}

@inproceedings{bayesLIME,
author = {Slack, Dylan and Hilgard, Sophie and Singh, Sameer and Lakkaraju, Himabindu},
title = {Reliable post hoc explanations: modeling uncertainty in explainability},
year = {2021},
isbn = {9781713845393},
publisher = {Curran Associates Inc.},
address = {Red Hook, NY, USA},
articleno = {719},
numpages = {14},
series = {NIPS '21}
}

@article{friedman1991multivariate,
  title={Multivariate adaptive regression splines},
  author={Friedman, Jerome H},
  journal={The annals of statistics},
  volume={19},
  number={1},
  pages={1--67},
  year={1991},
  publisher={Institute of Mathematical Statistics}
}

@BOOK{Fishman,
  title     = "Monte Carlo",
  author    = "Fishman, George",
  publisher = "Springer",
  series    = "Springer Series in Operations Research and Financial Engineering",
  month     =  may,
  year      =  2011,
  address   = "New York, NY",
  language  = "en"
}

@article{harman,
title = {On decompositional algorithms for uniform sampling from n-spheres and n-balls},
journal = {Journal of Multivariate Analysis},
volume = {101},
number = {10},
pages = {2297-2304},
year = {2010},
issn = {0047-259X},
doi = {https://doi.org/10.1016/j.jmva.2010.06.002},
author = {Radoslav Harman and Vladimír Lacko},
}

@inproceedings{lime,
author = {Ribeiro, Marco Tulio and Singh, Sameer and Guestrin, Carlos},
title = {"Why Should I Trust You?": Explaining the Predictions of Any Classifier},
year = {2016},
isbn = {9781450342322},
publisher = {Association for Computing Machinery},
address = {New York, NY, USA},
url = {https://doi.org/10.1145/2939672.2939778},
doi = {10.1145/2939672.2939778},
booktitle = {Proceedings of the 22nd ACM SIGKDD International Conference on Knowledge Discovery and Data Mining},
pages = {1135–1144},
numpages = {10},
location = {San Francisco, California, USA},
series = {KDD '16}
}

@article{diabetes,
   title={Least angle regression},
   volume={32},
   ISSN={0090-5364},
   url={http://dx.doi.org/10.1214/009053604000000067},
   DOI={10.1214/009053604000000067},
   number={2},
   journal={The Annals of Statistics},
   publisher={Institute of Mathematical Statistics},
   author={Efron, Bradley and Hastie, Trevor and Johnstone, Iain and Tibshirani, Robert},
   year={2004},
   month=apr }

@InProceedings{lime_survey,
author="Knab, Patrick
and Marton, Sascha
and Schlegel, Udo
and Bartelt, Christian",
editor="Guidotti, Riccardo
and Schmid, Ute
and Longo, Luca",
title="Which LIME Should I Trust? Concepts, Challenges, and Solutions",
booktitle="Explainable Artificial Intelligence",
year="2026",
publisher="Springer Nature Switzerland",
address="Cham",
pages="28--52",
isbn="978-3-032-08324-1"
}

@article{xai_survey,
title = {Explainable Artificial Intelligence (XAI): Concepts, taxonomies, opportunities and challenges toward responsible AI},
journal = {Information Fusion},
volume = {58},
pages = {82-115},
year = {2020},
issn = {1566-2535},
doi = {https://doi.org/10.1016/j.inffus.2019.12.012},
url = {https://www.sciencedirect.com/science/article/pii/S1566253519308103},
author = {Alejandro {Barredo Arrieta} and Natalia Díaz-Rodríguez and Javier {Del Ser} and Adrien Bennetot and Siham Tabik and Alberto Barbado and Salvador Garcia and Sergio Gil-Lopez and Daniel Molina and Richard Benjamins and Raja Chatila and Francisco Herrera},
}

\end{document}